\documentclass[11pt]{article}
\usepackage[T1]{fontenc}
\usepackage[utf8]{inputenc}
\usepackage{lmodern}
\usepackage{amsmath,amssymb}
\usepackage{graphicx}
\usepackage{booktabs,longtable,array,calc}
\usepackage{cite}
\usepackage{xurl}
\usepackage[hidelinks]{hyperref}
\usepackage{microtype}
\usepackage[margin=1in]{geometry}
\makeatletter
\def\maxwidth{\ifdim\Gin@nat@width>\linewidth\linewidth\else\Gin@nat@width\fi}
\def\maxheight{\ifdim\Gin@nat@height>0.78\textheight 0.78\textheight\else\Gin@nat@height\fi}
\makeatother
\setkeys{Gin}{width=\maxwidth,height=\maxheight,keepaspectratio}
\title{\textbf{RAGWarrant}\\[0.4em]
\large Evidence-Preserving Governance for RAG Policy Promotion\\
Under Quality, Cost, Latency, and Risk Constraints}
\author{Richard Krueger \and Lucas Krause \and Zach Pocquette}
\date{Independent open-source research project\\September 2026\\Preprint version 0.1.1-rc1}

\begin{document}
\maketitle

\begin{abstract}
Retrieval-augmented generation systems are extensively instrumented with metrics, benchmarks, traces, and automated judges, but these tools do not decide whether a proposed policy change is safe to release. We present RAGWarrant, an open-source promotion-control framework that treats deployment as a constrained evidence decision rather than a leaderboard choice. RAGWarrant normalizes evaluator outputs and operational telemetry, applies predeclared quality and hard-risk gates, assigns evidence-class claim ceilings, preserves negative outcomes, and emits auditable PROMOTE, BLOCK, REJECT, or INCONCLUSIVE decisions. We evaluate the framework across T2-RAGBench, MultiHop-RAG, CRAG, HotpotQA, synthetic reproduction, and bounded local generative experiments. On HotpotQA, operational savings were blocked because answer quality fell beyond the declared margin. A bounded CRAG study selected a lower-cost quality-tied policy, but related generative gains were unstable and a held-out guardrail failed closed. We claim an auditable promotion-control abstraction, not optimizer superiority, human validation, or production readiness. The tagged artifact reproduces from a fresh clone, runs as a hardened Docker job, accepts external evaluator exports, and verifies artifact integrity.
\end{abstract}


\medskip
\noindent\textbf{Code and artifacts:} \url{https://github.com/RAGWarrant/ragwarrant-governance}

\noindent\textbf{Release candidate:} v0.1.1-rc1 \quad \textbf{Commit:} \texttt{ff529005bf7d00a0c3f79ba991563f3923d63205}

\section{Introduction}\label{sec:introduction}

Retrieval-augmented generation (RAG) has become a standard architecture
for connecting language models to current, domain-specific, and
attributable evidence \cite{lewis2020rag,gao2023survey}. Yet a RAG system is not a single
model. Its behavior depends on document preparation, retrieval depth,
query routing, reranking, context assembly, generation, abstention, and
tool-use policies. Every proposed change is therefore multi-objective: a
policy may be faster but less complete, cheaper but less faithful,
stronger on average but fragile on a protected slice, or attractive on
validation data yet worse for a held-out population.

The evaluation ecosystem has advanced rapidly. RAGAS, ARES, RAGChecker,
RAGBench, CRAG, HotpotQA, and related benchmarks provide reference-free
or labeled metrics, fine-grained diagnostics, and structured evidence
\cite{es2023ragas,saadfalcon2023ares,ru2024ragchecker,friel2024ragbench,yang2024crag,yang2018hotpotqa,chen2024benchmarking}. Observability platforms capture traces, cost, latency,
and feedback \cite{langsmith2026,phoenix2026}, while optimization systems such as DSPy
search prompts, programs, or pipeline configurations against a chosen
metric \cite{khattab2023dspy}. These capabilities answer important measurement and
search questions, but they do not settle the release question: given
several plausible policies and incomplete, heterogeneous evidence, which
policy may replace the baseline, which should be rejected, and when
should the system refuse to decide?

RAGWarrant addresses that decision layer. Its central abstraction is not
a new answer-quality metric or a universal optimizer. It is an
evidence-preserving promotion controller. RAGWarrant ingests native
metrics or normalized evaluator exports, binds them to provenance and
evidence class, applies non-compensatory gates and predeclared decision
rules, emits a machine-readable decision, and preserves supporting and
contradicting evidence as an auditable bundle. A lower-cost or
lower-latency policy is promotable only when its quality evidence, split
integrity, security posture, provenance, and claim scope also pass.

RAGWarrant is intended to govern RAG policy promotion, not to discover,
certify, or imply a globally optimal RAG configuration.

The research question is whether a lightweight, open-source governance
layer can convert heterogeneous RAG metrics and telemetry into
reproducible promotion decisions while reducing false promotion of
policies that save cost or latency at unacceptable quality risk.

The distinction matters because deployment authority is often implicit.
In a research notebook, the highest validation score may become the de
facto recommendation; in an enterprise workflow, a cost dashboard or
latency trace may play the same role. RAGWarrant makes that authority
explicit and inspectable. The framework does not prevent teams from
choosing aggressive operating points, but it requires the quality floor,
hard disqualifiers, evidence scope, and decision reason to be recorded
before the result is presented as promotable.

This paper makes four contributions. First, it defines a
promotion-control formalism that separates metric generation from
release decisions and uses noninferiority-style quality gates to
evaluate operational gains. Second, it introduces an evidence ladder and
claim ceiling so fixture, development, confirmatory, generative, and
deployment evidence cannot be silently relabeled. Third, it
operationalizes fail-closed governance through machine-readable
decisions, an append-only evidence ledger, artifact-integrity checks,
and a validator that binds public claims to result classes. Fourth, it
releases and evaluates a reproducible implementation with public-mini
execution, external-evaluator adapters, selector ablations, bounded
public and generative studies, fresh-clone reproduction, and a hardened
Docker job.

\begin{figure}[tbp]
\centering
\includegraphics[width=0.98\linewidth]{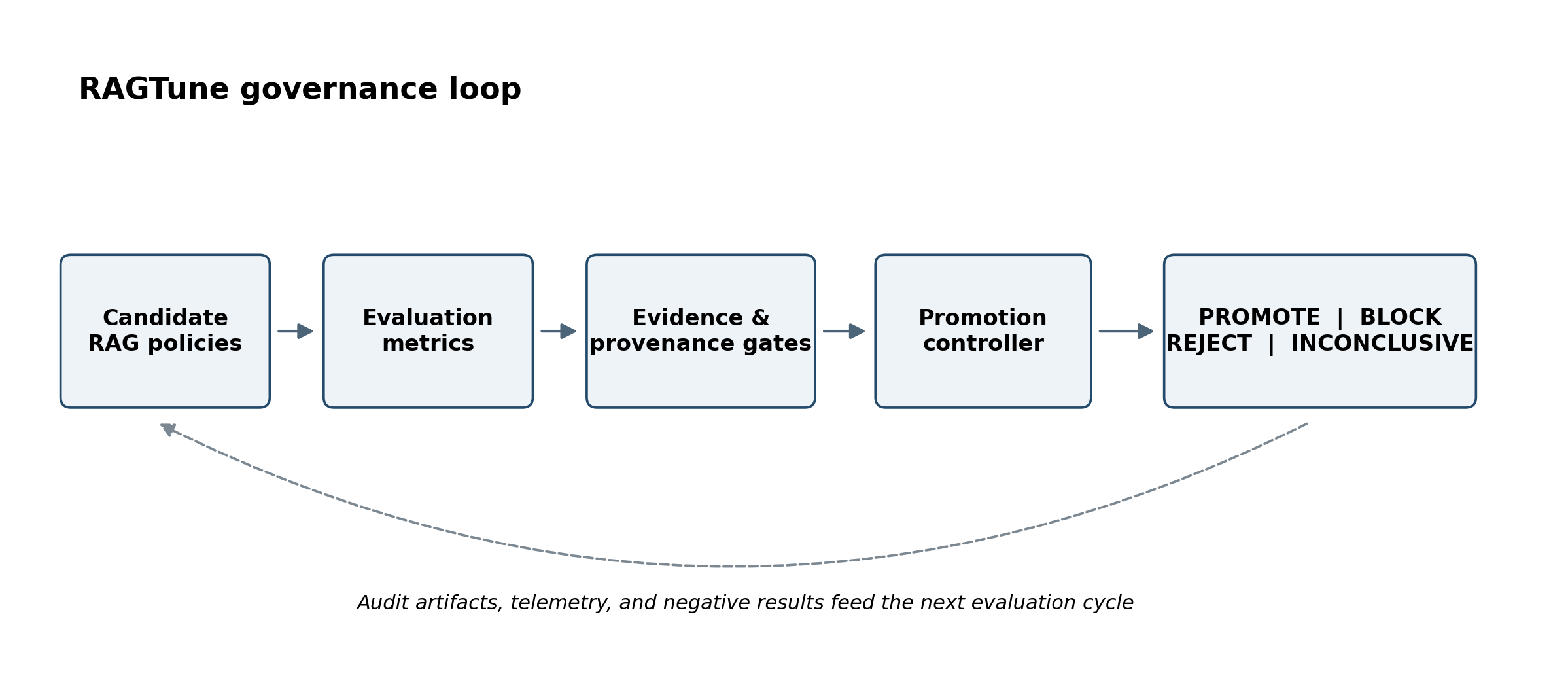}
\caption{RAGWarrant converts candidate-policy evidence into an auditable promotion decision; artifacts, telemetry, and negative results feed the next evaluation cycle.}
\label{fig:governance-loop}
\end{figure}

\section{Background and Related Work}\label{sec:background}

The original RAG formulation combined parametric generation with
non-parametric retrieval for knowledge-intensive tasks \cite{lewis2020rag}; later
work broadened the design space to modular retrieval, routing,
reranking, verification, and tool use \cite{gao2023survey}. This modularity creates a
policy-selection problem as well as an answer-generation problem.

RAGAS, ARES, RAGChecker, and RAGBench provide complementary evaluation
methods, from reference-free metrics to prediction-powered judges,
component diagnostics, and labeled benchmark data \cite{es2023ragas,saadfalcon2023ares,ru2024ragchecker,friel2024ragbench}. CRAG
stresses factuality, dynamism, long-tail entities, and mock web or
knowledge-graph APIs, while HotpotQA provides multi-hop questions and
sentence-level supporting facts \cite{yang2024crag,yang2018hotpotqa}. Benchmarking studies
likewise compare model and retrieval choices in RAG settings \cite{chen2024benchmarking}.
RAGWarrant consumes these measurements as bounded evidence; it does not
replace their metric semantics.

DSPy and related optimizers search candidate programs or policies
\cite{khattab2023dspy}, while LangSmith and Phoenix organize experiments, traces,
evaluator scores, latency, and feedback \cite{langsmith2026,phoenix2026}. The distinction is
concise: observability explains, evaluation estimates, optimization
searches, and governance decides. The broader ML-systems literature
similarly emphasizes hidden dependencies, testing, monitoring, and
technical debt beyond model code \cite{sculley2015debt,breck2017mltest}.

NIST's AI Risk Management Framework and Generative AI
Profile organize lifecycle governance around Govern, Map, Measure, and
Manage, and ISO/IEC 42001 specifies an AI management system \cite{tabassi2023nist,autio2024genai,iso42001}.
RAGWarrant is narrower and executable: it sits between measurement and
deployment and produces a traceable RAG-policy release decision.

\begin{longtable}[]{@{}
  >{\raggedright\arraybackslash}p{(\columnwidth - 6\tabcolsep) * \real{0.1929}}
  >{\raggedright\arraybackslash}p{(\columnwidth - 6\tabcolsep) * \real{0.2357}}
  >{\raggedright\arraybackslash}p{(\columnwidth - 6\tabcolsep) * \real{0.2500}}
  >{\raggedright\arraybackslash}p{(\columnwidth - 6\tabcolsep) * \real{0.3214}}@{}}
\caption{RAGWarrant complements metric, benchmark, optimization, observability, and management-framework layers.}\\

\toprule\noalign{}
\begin{minipage}[b]{\linewidth}\raggedright
System family
\end{minipage} & \begin{minipage}[b]{\linewidth}\raggedright
Primary emphasis
\end{minipage} & \begin{minipage}[b]{\linewidth}\raggedright
Typical output
\end{minipage} & \begin{minipage}[b]{\linewidth}\raggedright
RAGWarrant relationship
\end{minipage} \\
\midrule\noalign{}
\endhead
\bottomrule\noalign{}
\endlastfoot
RAGAS / ARES / RAGChecker & Quality and diagnostic evaluation & Metric
scores and component diagnoses & Consumes declared metrics as governance
evidence \\
RAGBench / CRAG / HotpotQA & Benchmark data and labels & Examples,
references, supporting facts, scores & Uses bounded evidence sources
with dataset-specific claim ceilings \\
DSPy and optimizers & Search over programs or policies & Candidate
maximizing an objective & Treats optimizer output as a candidate, not
automatic promotion \\
LangSmith / Phoenix & Experiments, traces, observability & Runs,
evaluator scores, operational telemetry & Normalizes exports and applies
release gates \\
NIST AI RMF / ISO 42001 & Lifecycle and management governance &
Practices, outcomes, management requirements & Implements a narrow
technical promotion control \\
RAGWarrant & Evidence-preserving promotion control & PROMOTE, BLOCK,
REJECT, or INCONCLUSIVE plus audit bundle & Coordinates evidence, gates,
claim ceilings, and integrity \\
\end{longtable}

\section{Design Goals and Threat Model}\label{sec:threat-model}

RAGWarrant is designed around false promotion: releasing a candidate
because it appears cheaper, faster, or better on an aggregate score when
its quality evidence is insufficient, a protected slice regresses,
provenance is invalid, or the result exceeds its evidence class. False
refusal also matters, but the current design deliberately prefers
conservative refusal when the cost of unnoticed quality loss is high.

The threat model assumes heterogeneous and imperfect evaluators. Each
suite therefore declares metric direction, weight, split structure,
evidence class, and hard gates rather than treating any single score as
ground truth.

\begin{itemize}
\item
  Metric risk: evaluator scores may be noisy, constant, weakly
  calibrated, or insensitive to meaningful policy differences.
\item
  Selection risk: validation winners may reverse on held-out data, and
  optimizers may overfit small development sets.
\item
  Operational risk: cost-only or latency-only selection can improve
  efficiency while degrading answer quality or evidence support.
\item
  Evidence risk: fixtures, development runs, frozen observations, and
  confirmatory runs support different claim ceilings.
\item
  Reproducibility risk: results may depend on mutable code, local state,
  unpinned data, or unverifiable artifacts.
\item
  Publication and security risk: licensed text, prompts, answers,
  credentials, private paths, or unsupported claims may leak into public
  artifacts.
\end{itemize}

The corresponding design objectives are to freeze decision criteria
before confirmatory evaluation; separate eligibility from ranking; make
protected regressions and security failures non-tradable; preserve
blocked and negative outcomes; trace every public claim to a result
artifact; remain agnostic to the evaluator that produced a metric; and
make the complete decision runnable as a finite, containerized job.
These objectives optimize for auditability and correction, not maximum
promotion throughput.

\section{RAGWarrant System}\label{sec:system}

\subsection{Policy candidates and normalized evidence}\label{sec:normalized-evidence}

A RAG policy is a named, versioned configuration that may change
retrieval depth, endpoints, routing, reranking, context limits, fallback
behavior, generator choice, abstention, or selector logic. Policies may
be hand-authored, proposed by an optimizer, or imported from another
experiment. RAGWarrant records behavioral parameters alongside
normalized evidence such as answer correctness, faithfulness, evidence
support, context recall, abstention correctness, API calls, cost,
latency, failure rate, and protected-regression indicators.

External-evaluator adapters map Ragas-like, DeepEval-like,
LangSmith-like, Phoenix-like, generic CSV, and generic JSONL exports
into a canonical schema while retaining evaluator provenance. The
release-candidate demonstration uses sanitized synthetic-shaped exports;
it establishes schema interoperability, not official integration. This
separation allows upstream metrics to evolve without changing the
downstream release contract.

Normalization is intentionally explicit. Each imported metric retains
its evaluator name, group, direction, scale, split, policy ID, and
source-artifact hash. Scores are not treated as interchangeable merely
because they share a 0-to-1 range. A suite may combine them into a
declared composite, use them as separate hard gates, or retain them only
for diagnosis. Operational telemetry follows the same principle: modeled
cost, measured latency, API-call count, and context volume remain
distinct fields so that a favorable weighted utility can be decomposed
during review.

A governance suite is configured as a declarative contract. It
identifies candidate and baseline policies, development and held-out
split roles, metric direction and composition, the noninferiority
margin, operational objectives, hard disqualifiers, protected slices,
evidence class, and expected artifacts. Candidate generation may be
exploratory, but selection logic is frozen before a confirmatory split
is evaluated. Confirmatory data estimate the consequences of the frozen
decision; they are not used to redesign the selector after the fact.

\subsection{Canonical promotion formalism}\label{sec:promotion-formalism}

Let $p$ be a candidate and $b$ the current baseline. Define quality and
operational deltas as $\Delta_q=q(p)-q(b)$, $\Delta_c=c(p)-c(b)$, and
$\Delta_l=l(p)-l(b)$, with positive quality better and negative cost or
latency better. A suite declares a quality-loss margin $\delta$ before
confirmatory evaluation. In the canonical rule, a candidate is
quality-noninferior when the lower confidence bound on $\Delta_q$ is no
worse than $-\delta$; an operational claim additionally requires the
relevant upper confidence bound on $\Delta_c$ or $\Delta_l$ to be below
zero. Promotion is permitted only if all hard gates pass.

\begin{align}
\Delta_q &= q(p)-q(b), \\
\Delta_c &= c(p)-c(b), \\
\Delta_l &= l(p)-l(b).
\end{align}

\begin{equation}
\operatorname{PROMOTE}(p) \iff \operatorname{hard\_gates}(p)
\land \operatorname{LCB}(\Delta_q) \ge -\delta
\land \left[\operatorname{UCB}(\Delta_c)<0 \;\lor\; \operatorname{UCB}(\Delta_l)<0\right].
\end{equation}

This use of noninferiority is procedural rather than clinical: it does
not make RAG evaluation equivalent to a randomized trial \cite{wellek2010equivalence,piaggio2012noninferiority}.
Its value is that acceptable loss is declared in advance, evidence
direction is explicit, and efficiency is not equated with release
eligibility. If uncertainty crosses a gate, the decision is INCONCLUSIVE
rather than a win.

\begin{figure}[tbp]
\centering
\includegraphics[width=0.62\linewidth]{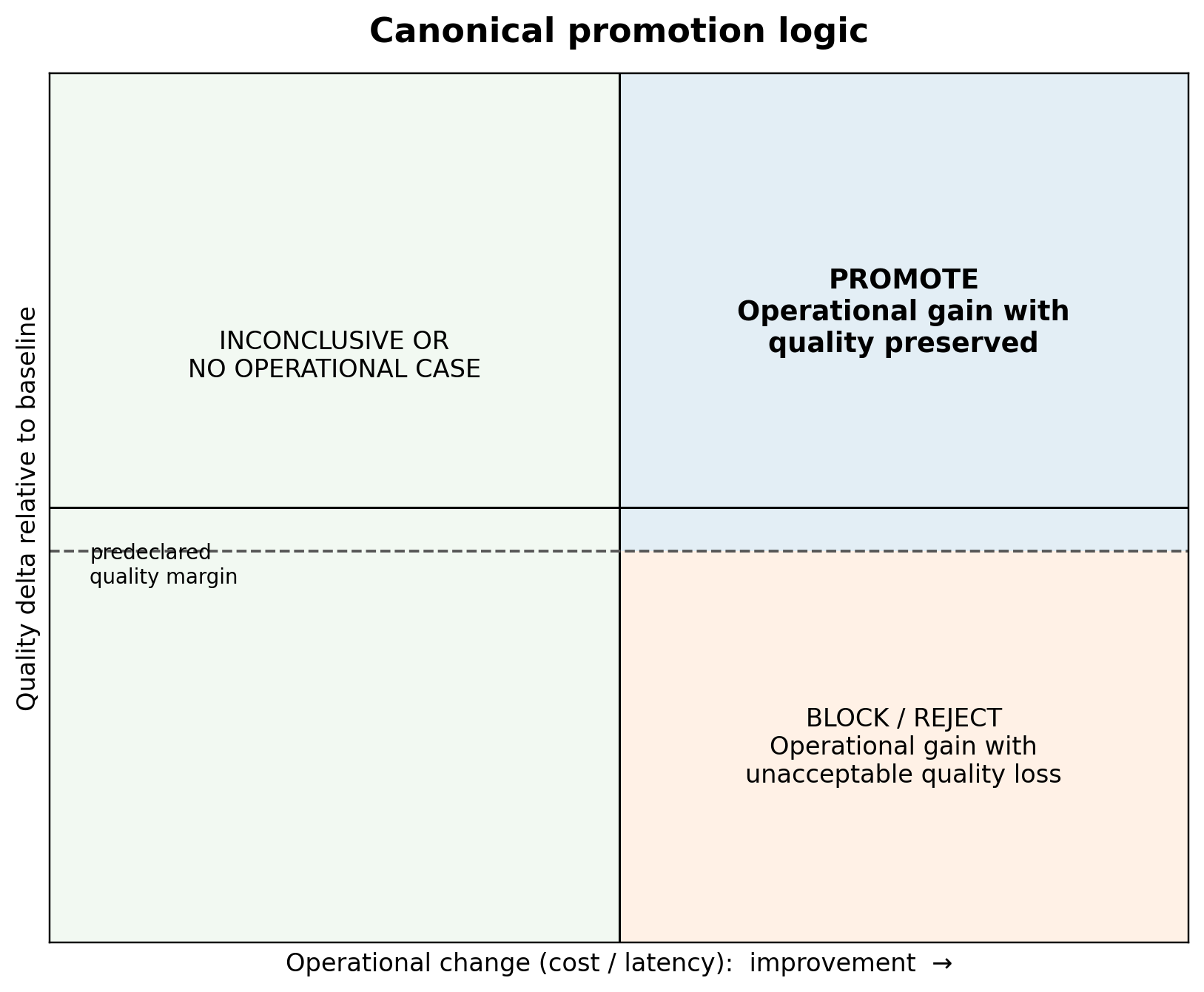}
\caption{Canonical decision geometry. Operational improvement is promotable only when the quality-preservation condition and all hard gates pass.}
\label{fig:decision-geometry}
\end{figure}

\subsection{Hard gates and decision taxonomy}\label{sec:decision-taxonomy}

Hard gates are non-compensatory. Leakage, missing provenance, security
disqualification, publication-hygiene failure, or an absent quality
signal cannot be offset by lower cost. Weighted utilities can rank
eligible candidates, but eligibility is established first. The
deployable job exposes four scientific decisions: PROMOTE when evidence
supports replacement within scope; BLOCK when the decision cannot be
evaluated safely; REJECT when evidence favors retaining the baseline;
and INCONCLUSIVE when available evidence does not resolve the gate.
Runtime errors are recorded separately in the machine-readable schema;
the full taxonomy appears in Appendix A.

\begin{longtable}[]{@{}
  >{\raggedright\arraybackslash}p{(\columnwidth - 6\tabcolsep) * \real{0.2100}}
  >{\raggedright\arraybackslash}p{(\columnwidth - 6\tabcolsep) * \real{0.2250}}
  >{\raggedright\arraybackslash}p{(\columnwidth - 6\tabcolsep) * \real{0.3150}}
  >{\raggedright\arraybackslash}p{(\columnwidth - 6\tabcolsep) * \real{0.2500}}@{}}
\caption{Scientific decision classes used by the deployable governance job; runtime errors are represented separately in the schema.}\\

\toprule\noalign{}
\begin{minipage}[b]{\linewidth}\raggedright
Decision
\end{minipage} & \begin{minipage}[b]{\linewidth}\raggedright
Meaning
\end{minipage} & \begin{minipage}[b]{\linewidth}\raggedright
Typical trigger
\end{minipage} & \begin{minipage}[b]{\linewidth}\raggedright
Consequence
\end{minipage} \\
\midrule\noalign{}
\endhead
\bottomrule\noalign{}
\endlastfoot
PROMOTE & Evidence supports replacing the baseline within declared
scope. & Quality and operational/risk objectives pass; no hard
disqualifier. & Candidate may advance with recorded boundaries. \\
BLOCK & Promotion cannot be evaluated safely. & No usable signal,
leakage, missing provenance, security or hygiene failure. & No release
claim; blocker and remediation are recorded. \\
REJECT & Evidence favors retaining the baseline. & Confirmed quality
loss, negative held-out result, or protected regression. & Candidate is
not promoted; negative evidence remains append-only. \\
INCONCLUSIVE & Available evidence does not resolve the decision. &
Interval crosses threshold, evidence is mixed, or sample is
underpowered. & No promotion; additional evidence may be collected. \\
\end{longtable}

\subsection{Evidence classes and claim ceilings}\label{sec:evidence-classes}

Every run is assigned an evidence class. Fixtures establish execution
and schema behavior, not policy superiority. Development runs guide
design but cannot be relabeled confirmatory. Frozen observations support
ablation but are weaker than independent collection. Generative evidence
requires a pinned generator and usable answer-quality signal; human or
platform claims require the corresponding annotations or platform-native
records. The evidence ladder therefore sets a claim ceiling: evidence
may support a narrower statement than its maximum, but never a broader
one.

\begin{figure}[tbp]
\centering
\includegraphics[width=0.72\linewidth]{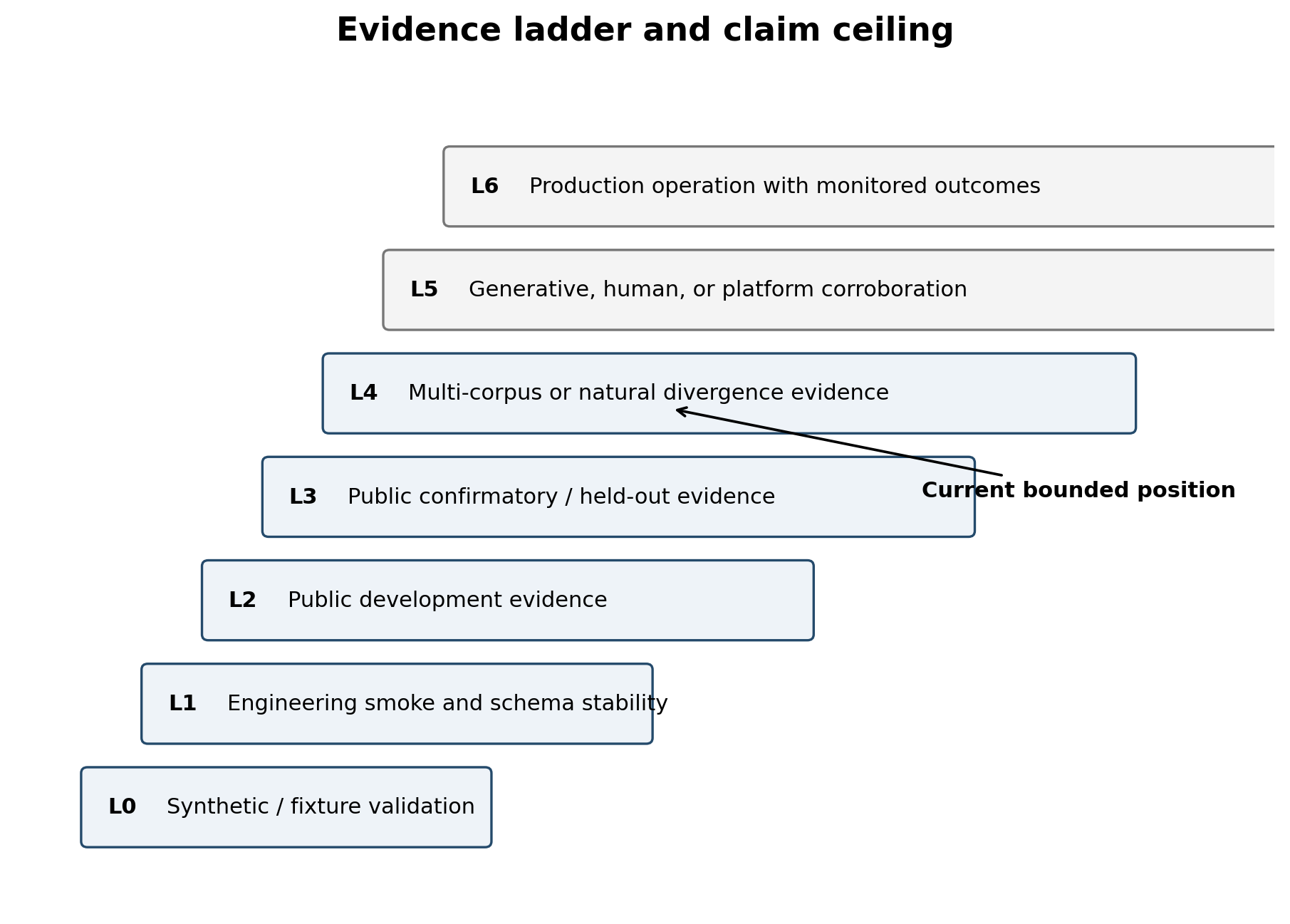}
\caption{Evidence classes define a claim ceiling. The release reaches bounded public and generative evidence in selected settings, not the highest evidence levels.}
\label{fig:evidence-ladder}
\end{figure}

\subsection{Claim validation, integrity, and deployment contract}\label{sec:claim-validation}

The publication validator checks that public language and artifacts
match those ceilings. It scans for raw licensed text, prompts, answers,
API responses, credentials, private paths, large tracked files, and
unsupported claim patterns. A historical ledger preserves positive,
negative, blocked, refused, and superseded outcomes, while verify-run
checks required artifacts, schemas, and hashes. Together, these
mechanisms prevent later summaries from reviving a claim that a stronger
audit or held-out test had already narrowed.

Claim validation occurs after scientific decision generation but before
publication or promotion artifacts are accepted. The validator compares
result classes with allowed language, confirms required evidence files,
and checks that high-risk fields appear only as hashes, counts, or
sanitized labels. This creates two independent failure channels: a
candidate can fail the scientific gate, or an otherwise valid run can
fail artifact hygiene. Neither failure is converted into evidence for
the candidate.

The finite job contract accepts policy configurations, sanitized
datasets or evaluator exports, and operational telemetry; it emits a
promotion decision, run manifest, policy summaries, selector comparison,
claim update, validation report, and optional audit bundle. The same OCI
image is packaged for local Docker, Docker Compose, Kubernetes Jobs, and
cloud job templates. Only local hardened Docker execution is
demonstrated here; the architecture is portable, but the evidence claim
remains local deployment validation.

\begin{figure}[tbp]
\centering
\includegraphics[width=0.94\linewidth]{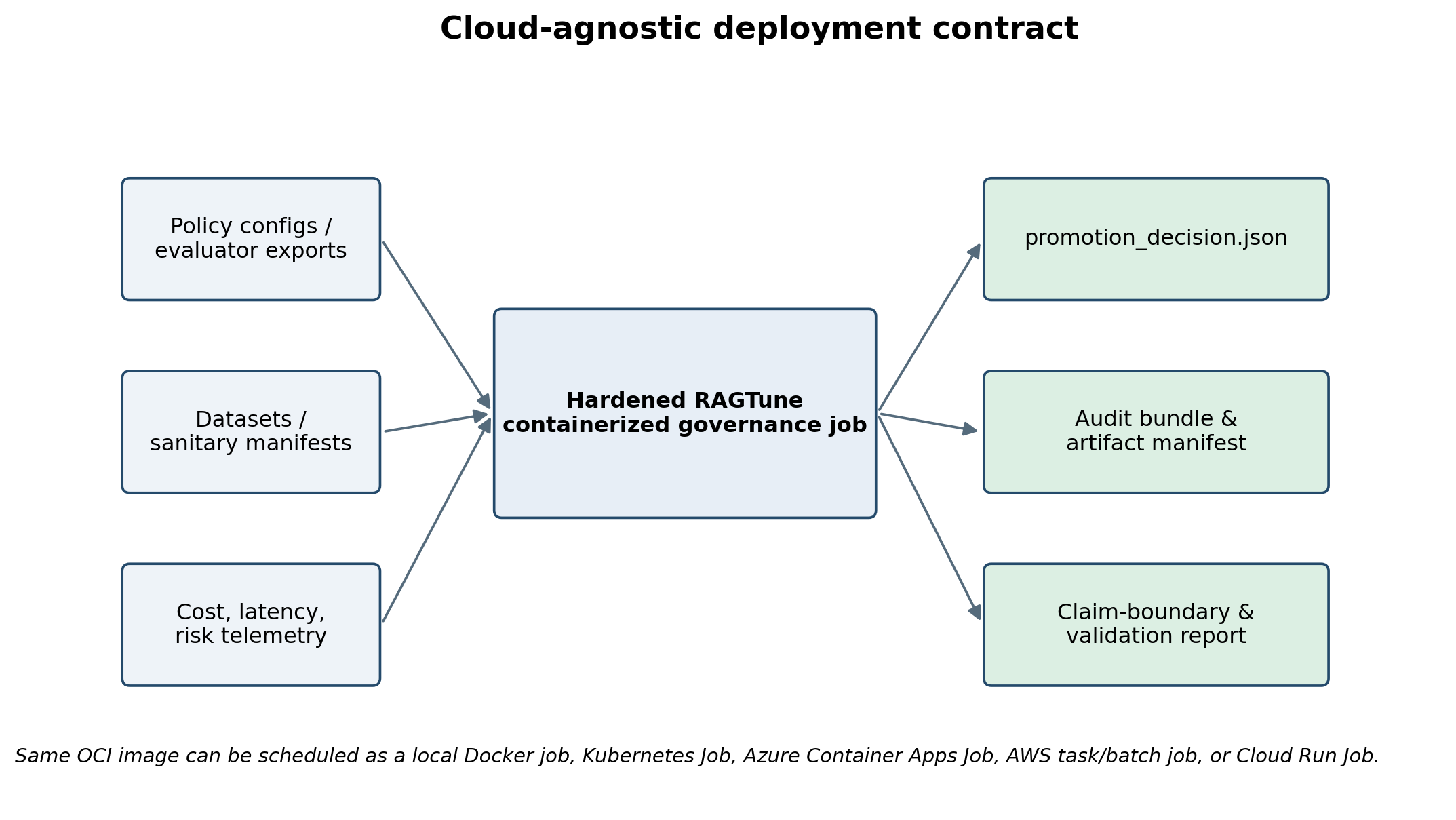}
\caption{Cloud-agnostic job contract. A finite container consumes policies and evidence, then emits a promotion decision, audit bundle, and validation report.}
\label{fig:deployment-contract}
\end{figure}

\section{Experimental Program}\label{sec:experimental-program}

The empirical program progressed from fixture-level orchestration to
public development and confirmatory corpora, statistical audits,
corpus-backed retrieval, mock APIs, generated-answer evaluation,
behaviorally distinct policies, held-out offsets, selector ablations,
and deployment validation. It is a longitudinal research-and-engineering
program rather than one preregistered trial. Claims are therefore
reported by evidence class, and Table 3 identifies each
corpus\textquotesingle s role and principal limitation.

Dataset use was constrained by licensing and by whether policies could
actually change retrieval. T2-RAGBench \cite{strich2026t2ragbench} supplied a larger
text-and-table development path; MultiHop-RAG \cite{tang2024multihoprag} provided a fresh
sealed multi-hop confirmatory corpus; CRAG enabled both corpus-backed
retrieval and a mock-API routing environment; and HotpotQA added answer
labels and supporting-fact annotations. The public repository carries
manifests, hashes, and processed metrics rather than raw licensed text.
Split construction used deterministic grouping and duplicate checks
where supported, and confirmatory unlocks were refused when provenance,
leakage, or evidence prerequisites failed.

\begin{longtable}[]{@{}
  >{\raggedright\arraybackslash}p{(\columnwidth - 6\tabcolsep) * \real{0.1643}}
  >{\raggedright\arraybackslash}p{(\columnwidth - 6\tabcolsep) * \real{0.2071}}
  >{\raggedright\arraybackslash}p{(\columnwidth - 6\tabcolsep) * \real{0.3071}}
  >{\raggedright\arraybackslash}p{(\columnwidth - 6\tabcolsep) * \real{0.3214}}@{}}
\caption{Principal datasets and evidence roles. Raw licensed datasets are not redistributed by the public repository.}\\

\toprule\noalign{}
\begin{minipage}[b]{\linewidth}\raggedright
Dataset / path
\end{minipage} & \begin{minipage}[b]{\linewidth}\raggedright
Role
\end{minipage} & \begin{minipage}[b]{\linewidth}\raggedright
Evidence characteristics
\end{minipage} & \begin{minipage}[b]{\linewidth}\raggedright
Primary limitation
\end{minipage} \\
\midrule\noalign{}
\endhead
\bottomrule\noalign{}
\endlastfoot
T2-RAGBench & Public end-to-end development & 1,142-query development
corpus; behaviorally variable policies & Development evidence;
deterministic extractive generation in key runs \\
MultiHop-RAG & Public confirmatory corpus & 331-query sealed
confirmatory test; full corpus-backed path & Governance matched
quality-only \\
RAGBench HotpotQA & Context-retrieval enablement & Policy-dependent
context assembly & Not full source-corpus retrieval in packaged
evidence \\
CRAG web documents & Full corpus-backed retrieval & 2,706 rows; 9,848
web documents; 571 confirmatory rows & Noncommercial restriction;
governance matched quality-only \\
CRAG mock API & Tool routing and generative validation & Live and frozen
paths; calls, cost, latency, evaluator mapping & Strongest positive
result bounded; generative results unstable \\
HotpotQA & Alternate corpus with answer labels & 1,000 local examples;
249 confirmatory behavioral rows & Operational gain accompanied by
quality loss \\
Public mini & Open-source reproduction & Synthetic, deterministic, no
private data or model & Onboarding proof, not external science \\
\end{longtable}

Candidate policies included static defaults, low and expanded retrieval,
adaptive routing, cost- and latency-aware policies, greedy and
Optuna/TPE search, constrained optimization, Pareto selection, RAG
Compass, and governed or risk-guarded selectors. The policy under
evaluation is always a candidate, never an automatic beneficiary of the
framework.

Generative studies used pinned local models through Ollama and stored
prompts and answers outside the public tree. Public rows retained
hashes, lengths, policy identifiers, evaluator scores, and operational
telemetry. The program tested qwen3:8b, gpt-oss:20b, and llama3.2:3b in
bounded slices, not to rank the models, but to determine whether a
governance conclusion persisted when answer emission and model behavior
changed. Fixed offsets and cross-offset guardrails were introduced after
the first favorable slice to test stability rather than enlarge the same
result.

Where paired example-level evidence existed, analyses used paired
bootstrap intervals and win/tie/loss summaries. Grouped intervals were
reported only when separately implemented. Noninferiority margins, often
0.01 for a composite quality score, were declared in configuration.
Statistical output remained subordinate to evidence validity: a narrow
interval cannot rescue duplicated rows, a constant quality signal,
invalid provenance, or confirmatory leakage.

The primary decision unit was a policy comparison under a declared
baseline and scope. Quality-only selection used validation quality
without operational penalties; constrained and governed selectors
applied feasibility requirements; Pareto analysis exposed nondominated
policies without forcing a scalar utility; and oracle variants were
retained only as ceilings. When a quality signal was constant, missing,
or demonstrably duplicated, the run was blocked even if cost and latency
differed.

\section{Results}\label{sec:results}

Table 4 is the numerical center of the paper; the prose below interprets
rather than duplicates it. An earlier zero-width bootstrap episode also
shaped the program: row-level reconstruction showed that a seemingly
precise result repeated one aggregate paired delta rather than
calibrated query-level uncertainty. The result was retained but
downgraded, and later suites added explicit nonconstant-signal and
grouped-analysis checks. Appendix D records that audit history.

Representative artifact pointers for Table 4 are retained in the public
tree rather than reproduced as raw data. The strongest CRAG mock-API
result is associated with run ID \path{RAGWarrant_crag_mock_api_validation_v1_20260809-165415-92d8c0edd4};
supporting summaries include
\path{results/behavioral_governance/paper_ready_summary.md},
\path{results/multi_dataset_behavioral_governance/paper_ready_summary.md},
\path{results/generative_llm_validation/synthesis_report.md},
\path{docs/selector_ablation_stress_v2.md},
\path{results/claim_status/claim_status_table.csv}, and the canonical
\path{results/run_index.csv}.

The results should therefore be read as tests of a release mechanism,
not as a tournament among retrievers. A tie can establish that the
mechanism executes without adding value on a corpus; a negative result
can establish that the gate refuses an efficiency gain; and a failed
replication can narrow the claim supported by an earlier slice. The
evidence ledger makes those outcomes cumulative rather than disposable.

\begin{figure}[tbp]
\centering
\includegraphics[width=0.98\linewidth]{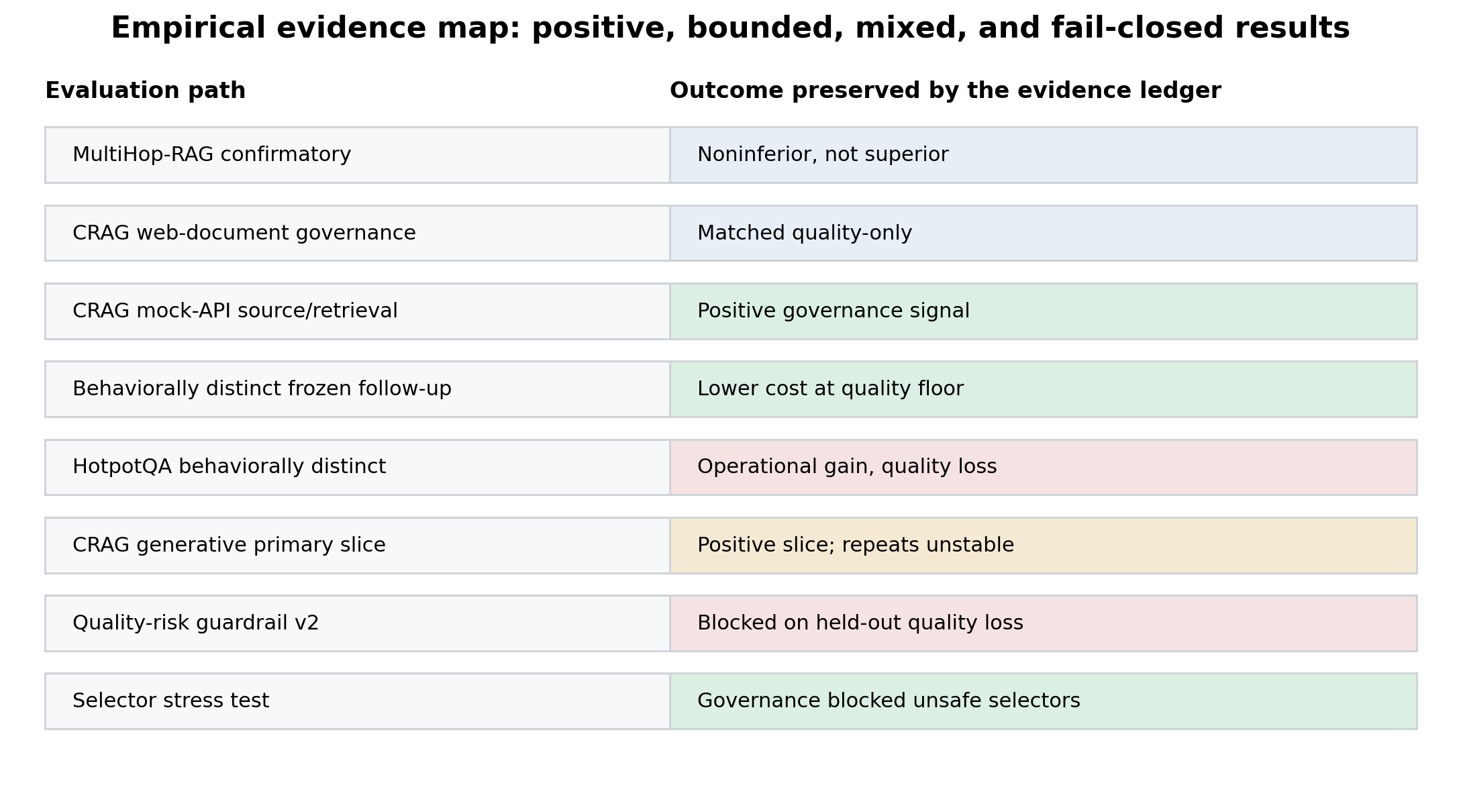}
\caption{The evidence ledger preserves positive, bounded, mixed, negative, and fail-closed outcomes instead of collapsing them into one leaderboard.}
\label{fig:evidence-map}
\end{figure}

\subsection{Confirmatory outcomes: noninferiority without general superiority}\label{sec:confirmatory}

Public confirmatory studies most often showed that governance was
feasible without establishing an advantage over quality-only selection.
On MultiHop-RAG, governed and quality-only aliases selected the same
policy, and an audit showed that a held-out Optuna/TPE win was a
validation-to-test reversal rather than a selector bug. On CRAG
web-document retrieval, both selectors again chose the same
high-retrieval policy. A larger T2-RAGBench development run was directly
unfavorable to RAG Compass, reinforcing the optimizer-agnostic design.

The strongest bounded source/retrieval signal came from the CRAG
mock-API path. Governance selected a lower-budget policy while
quality-only selection favored a regression-aware search policy. The
reported utility difference was approximately +0.001 and stable across
most cost-latency weight settings, but ablation attributed nearly all of
it to declared cost and a small latency difference, not raw-quality
improvement. It is therefore evidence that a promotion controller can
choose a cheaper quality-tied candidate, not that it discovered a
universally better retriever.

A behaviorally distinct follow-up gave the result a more operational
form: a single-endpoint policy used materially fewer calls and lower
cost than Optuna/TPE while remaining within the declared
proxy-plus-evidence quality margin. That comparison is stronger than a
cosmetic utility tie-break, yet it remains derived from frozen
observations rather than an independent live or human-calibrated
evaluation.

These studies also expose an important distinction between selection
correctness and hindsight regret. A policy selected from validation
evidence can lose to another policy on a sealed test set without
implying a logic defect. RAGWarrant records that reversal, but it does
not retroactively choose the test winner. This preserves the separation
between model selection and final estimation that the governance
contract is intended to enforce.

\subsection{Fail-closed outcomes}\label{sec:fail-closed}

HotpotQA provides the clearest negative test. The governed policy
reduced retrieval cost, context volume, and calls, but answer quality
declined beyond the declared noninferiority margin. A conventional
efficiency report could describe the run as a cost win; RAGWarrant
classified it as operational gain with quality loss and withheld
promotion.

A separate pooled quality-risk guardrail trained only on deployable
metadata and passed its validation gates across four folds. Held-out
evaluation then exposed quality-loss risk on three offsets. Although API
calls and cost fell, the strict gate blocked the selector. These two
cases are the empirical core of the governance claim: the
framework's value is visible not only when it identifies
an eligible candidate, but when it prevents attractive operational
deltas from becoming unsupported release decisions.

Fail-closed behavior is costly: a conservative controller can retain an
expensive baseline or delay a useful change. The packaged artifacts
therefore record blocked rate and promotion rate alongside quality
outcomes. The current program does not prove that the chosen trade-off
is optimal; it demonstrates that the trade-off is explicit,
reproducible, and reviewable.

\subsection{Generative feasibility without stability}\label{sec:generative}

Local generative evaluation became technically feasible after
answer-emission and evaluator-mapping repairs. A bounded CRAG slice
using a pinned qwen3:8b model produced non-empty, variable answers and
supported a cost-at-equivalent-generated-quality decision on its primary
offset. The point estimate was favorable and the policy reduced calls
and latency substantially.

The effect was not stable. Three deterministic repeats did not reproduce
the cost result; a second model produced usable signals without positive
cost slices; and a smaller instruct model repaired blank-answer behavior
without creating a repeatable efficiency win. HotpotQA likewise
confirmed nonconstant generated quality, answer diversity, and evidence
variance, but the audit covered only a small sample and did not
establish governance improvement. The generative contribution is
therefore machinery and bounded evidence, not a stable optimization
claim.

The model comparisons also revealed engineering failure modes that a
leaderboard would obscure. Thinking-mode output suppressed usable
answers in one path; another model produced high blank-answer rates; and
a smaller instruct model repaired emission without improving the
selection result. Treating generator availability, parse success, and
quality-signal variance as gates prevented these failures from being
mistaken for policy evidence.

\subsection{Selector ablation and systems results}\label{sec:selector-ablation}

The packaged selector stress test compared quality-only, cost-only,
latency-only, random, static, RAG Compass, governed noninferiority,
risk-guarded, and oracle-style selectors across sanitized evidence
families. Cost-only, latency-only, and random selection admitted
quality-loss cases; governed selectors traded promotion rate for zero
recorded quality-loss labels in the packaged cases. These rates are not
population estimates, but they show that non-compensatory gates create
observable behavior distinct from operational ranking.

A fresh Git clone installed and reproduced the public-mini workflow
without private data or model credentials; the hardened Docker job
completed under the constrained runtime profile; and verify-run
confirmed artifact hashes, schemas, and hygiene conditions. These
results establish that an independent reviewer can execute the contract
that produced a decision --- the public mini is deliberately synthetic
and fail-closed, proving installation, schemas, exit behavior, and claim
checks without implying external efficacy.

\begin{figure}[tbp]
\centering
\includegraphics[width=0.98\linewidth]{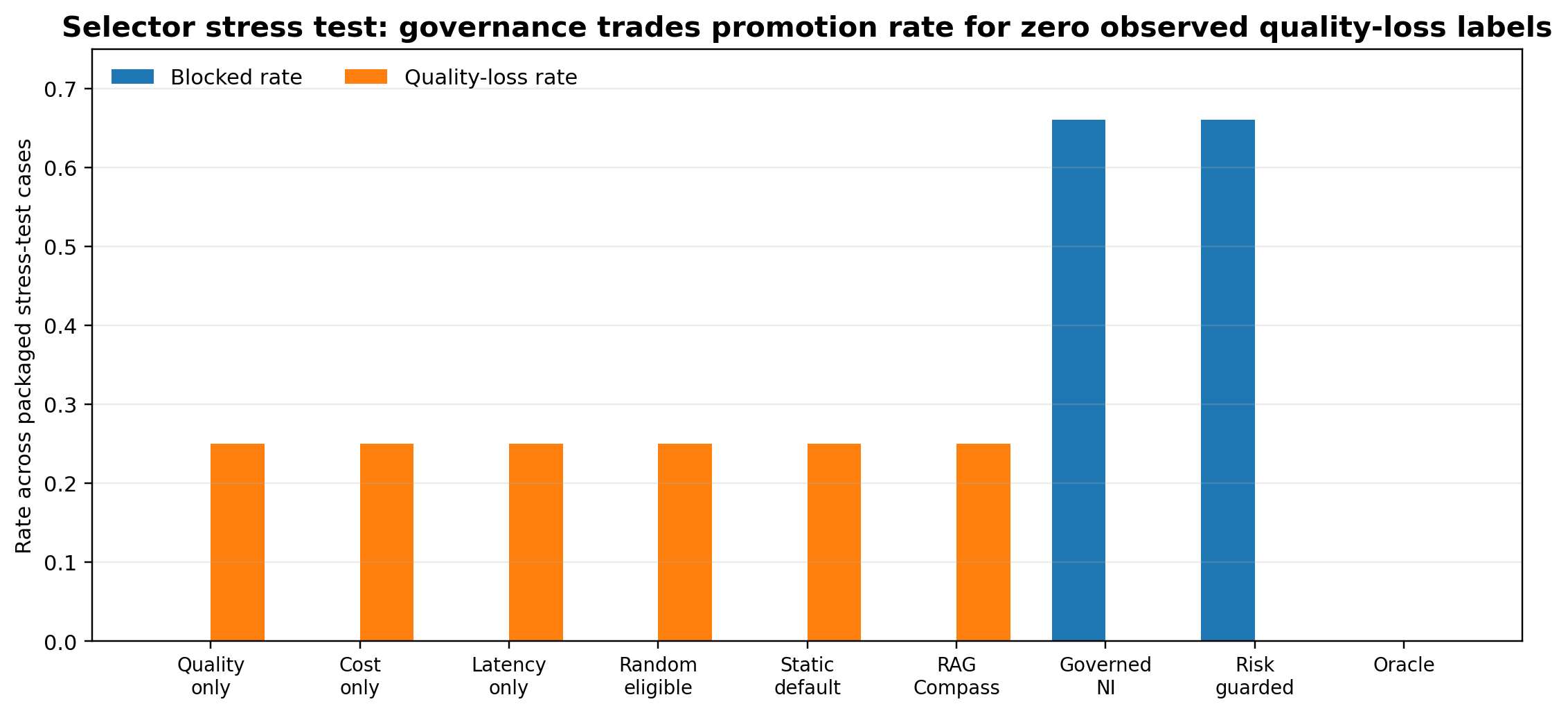}
\caption{In the packaged selector stress test, governed selectors blocked more cases and recorded no quality-loss labels; operational selectors did not block and admitted quality loss.}
\label{fig:selector-stress}
\end{figure}

\begin{longtable}[]{@{}
  >{\raggedright\arraybackslash}p{(\columnwidth - 6\tabcolsep) * \real{0.1786}}
  >{\raggedright\arraybackslash}p{(\columnwidth - 6\tabcolsep) * \real{0.1857}}
  >{\raggedright\arraybackslash}p{(\columnwidth - 6\tabcolsep) * \real{0.3071}}
  >{\raggedright\arraybackslash}p{(\columnwidth - 6\tabcolsep) * \real{0.3286}}@{}}
\caption{Principal empirical and systems results. Full precision is retained here; negative and mixed outcomes are part of the reported evidence.}\\

\toprule\noalign{}
\begin{minipage}[b]{\linewidth}\raggedright
Evaluation
\end{minipage} & \begin{minipage}[b]{\linewidth}\raggedright
Scope / N
\end{minipage} & \begin{minipage}[b]{\linewidth}\raggedright
Primary outcome
\end{minipage} & \begin{minipage}[b]{\linewidth}\raggedright
Interpretation
\end{minipage} \\
\midrule\noalign{}
\endhead
\bottomrule\noalign{}
\endlastfoot
MultiHop-RAG confirmatory & 331 queries & Governance delta 0;
noninferior, not superior & Valid confirmatory execution; no governance
advantage \\
CRAG web documents & 571 rows & Governed = quality-only = top\_k\_high &
Corpus-backed execution; no superiority \\
CRAG mock API & 571 rows & Utility +0.00100254; 571/0/0 wins/ties/losses
& Strongest bounded source/retrieval signal; largely cost/latency
driven \\
Frozen behaviorally distinct CRAG & 571 paired observations & Quality
-0.00515371; cost -2.36839; calls -1.79860 & Lower operating burden
within declared proxy-quality margin; derived evidence \\
HotpotQA behavioral & 249 confirmatory rows & Quality -0.0408367; F1
-0.0843373; cost -2.33014; calls -1.98394 & Operational gain with
quality loss; blocked \\
CRAG generative primary & 12 examples; 132 generations & Quality
+0.0166534; cost -3.77900; latency -5,971.85 ms & Positive bounded
slice; repeats/models did not reproduce \\
Guardrail v2 & 4 held-out offsets & 3 quality-loss blocks; 0 positive
latency results & Validation success failed held-out generalization \\
Selector stress & Packaged evidence families & Governed blocked 0.66
with 0 quality loss; cost/latency/random quality loss 0.25 & Gates
traded promotion rate for no observed quality-loss labels \\
Public mini + Docker & Synthetic deterministic job & Fresh clone,
hardened Docker, decision artifact, verify-run pass & Strong systems
reproducibility; not external empirical evidence \\
\end{longtable}

Note: cost and latency deltas in Table 4 follow each
suite\textquotesingle s declared telemetry fields and are comparable
only within the corresponding evaluation path. Unless an artifact
declares otherwise, cost is a normalized or modeled operational-cost
field rather than a dollar-denominated cloud bill.

\section{Discussion}\label{sec:discussion}

RAGWarrant's novelty is architectural and procedural
rather than metric-level. It separates evidence production from release
authority; applies quality preservation and hard gates before utility
ranking; ties each result to an evidence class and claim ceiling; and
packages the decision, contradictory evidence, and integrity checks as a
portable artifact. Noninferiority, risk management, observability, and
ML testing are established ideas \cite{tabassi2023nist,autio2024genai,iso42001,sculley2015debt,breck2017mltest,wellek2010equivalence,piaggio2012noninferiority}. The contribution is their
RAG-specific operationalization in a lightweight, vendor-neutral
promotion-control layer with an empirical record that includes failed
replications and refused promotions.

This position is complementary to existing tools. A team may obtain
answer correctness from one evaluator, faithfulness from another, cost
and latency from traces, and a security disqualifier from a separate
suite. RAGWarrant's role is to combine those declared
inputs under one release contract. The output is not another score but a
decision with reasons, limits, and traceable artifacts. This is
particularly useful when a quality-only winner, a constrained optimizer,
and a risk gate disagree.

The negative outcomes are consequently informative. HotpotQA and the
held-out guardrail show that lower cost or latency can coexist with
unacceptable answer degradation, while the unstable generative slices
show how quickly a positive local result can evaporate under offsets or
models. The current evidence does not estimate an optimal
false-promotion versus false-refusal trade-off; it establishes that the
framework can encode a conservative policy and preserve the cost of that
conservatism for later study.

A practical governance program must eventually price false refusal as
carefully as false promotion. Blocking too often can preserve a costly
configuration, slow improvement, and encourage teams to bypass the
process. RAGWarrant's current result taxonomy and
artifact schema make that future calibration possible because they
retain both the rejected candidate\textquotesingle s operational benefit
and the gate that prevented release. Larger independent studies can then
ask whether margins and risk thresholds produce an acceptable decision
curve rather than only whether any single policy wins.

The product contract is most relevant where changes require a defensible
release record: healthcare policy assistants, financial and insurance
knowledge systems, legal research, government services, life-sciences
evidence workflows, and internal enterprise copilots. In such settings
the question is rarely just which score is highest. It is whether a
change is justified for a declared corpus, use case, user population,
access tier, quality floor, latency target, and risk class. RAGWarrant
can run as a finite CI/CD or monitoring job and emit an artifact
suitable for change review; integration with live identity,
authorization, and incident processes remains deployment-specific.

The same abstraction can support layered enterprise policies. A release
contract may inherit organization-wide security and data-access
requirements, then specialize by corpus, use case, user population,
query class, and service-level objective. The optimizer may vary
retrieval depth or routing inside that envelope, but it cannot override
entitlement, source-authority, citation, or human-review rules. This
separation makes governance portable across customers without pretending
that one globally optimal RAG configuration exists.

\section{Limitations and Threats to Validity}\label{sec:limitations}

Empirical breadth remains limited. Few public corpora support full,
policy-dependent, corpus-backed evaluation in the packaged program; CRAG
also carries noncommercial-research restrictions. MultiHop-RAG and CRAG
web-document studies produced ties rather than superiority, and some
HotpotQA phases use context-retrieval evidence rather than a fully
reconstructed source corpus.

Generative samples are small and unstable. The favorable CRAG primary
slice contained 12 examples, repeats and alternative models did not
recover its cost result, and the larger HotpotQA target was not reached.
Composite proxy and automatic evaluator scores are not substitutes for
human judgment, and calibration across domains remains open.

The strongest positive source/retrieval result is also unusually small
in practical effect and heavily dependent on the declared cost model. It
should be interpreted as proof of decision behavior, not as a material
quality advance. The selector stress test is assembled from packaged
cases and cannot estimate real-world base rates of unsafe promotion.
Likewise, sanitized frozen-observation analyses are useful for ablation
but share dependencies with their parent runs.

The program is longitudinal and adaptive rather than a single
preregistered experiment. Repeated design changes create multiplicity
and researcher degrees of freedom; some observations are dependent, and
early intervals were demonstrably low-information. The response is
procedural rather than statistical sleight of hand: preserve chronology,
distinguish development from confirmatory evidence, report grouped
analyses only when implemented, and avoid a pooled significance claim.

Systems evidence is also bounded. Fresh-clone and hardened Docker
execution establish reproducibility of the public-mini contract, not
operational reliability under real authentication, authorization,
retention, availability, or incident-management requirements. External
evaluator adapters use synthetic-shaped exports rather than official
platform runs, and the security scan set is incomplete. RAG Compass
remains an optional candidate and has no supported superiority claim.

Reproducibility artifacts verify code paths, decisions, and public
hygiene; they cannot independently verify licensed raw data that are
excluded from the repository. Exact external replication therefore
depends on acquiring the original corpora and reconstructing the
declared manifests. The release tag itself is an engineering checkpoint
rather than a guarantee that every historical artifact can be
regenerated without those sources.

\section{Reproducibility, Availability, and Responsible Release}\label{sec:reproducibility}

Project code is released under Apache-2.0; third-party datasets retain
their original licenses, and raw CRAG or HotpotQA text is not
redistributed. The tagged v0.1.1-rc1 artifact identifies a fixed commit
and includes configs, schemas, sanitized processed results, tests,
Docker assets, deployment templates, claim tables, release manifests,
and an evidence ledger.

Public artifacts exclude raw licensed dataset text, raw prompts,
generated answers, raw API responses, credentials, local caches, and
model weights; where needed, the repository retains hashes, IDs, counts,
metrics, and sanitized summaries.

A fresh clone can run the deterministic public-mini job without private
data, a generator, or cloud credentials. The hardened container
reproduces the same governance contract, while verify-run checks the
artifact manifest, hashes, schemas, and hygiene conditions.
Dataset-dependent studies require users to acquire the original data
under their licenses. These mechanisms make the decision logic and
failure modes independently inspectable even when licensed raw evidence
cannot be published.

GitHub CI executes publication tests and claim checks on the public
tree. Deployment templates target Kubernetes and major cloud job
primitives, but they are intentionally separated from the scientific
claim: a template shows how to schedule the container, whereas platform
evidence would require a real platform-native run record.

\section{Conclusion}\label{sec:conclusion}

RAGWarrant makes release judgment executable. It normalizes evidence,
applies non-compensatory gates, uses predeclared quality margins where
appropriate, preserves negative outcomes, emits machine-readable
decisions, validates public claims, and packages the result as a
reproducible containerized job. The empirical record demonstrates the
property most important to this design: apparent efficiency gains can be
refused when quality, held-out, or evidence-validity checks do not
survive.

That boundary is the paper\textquotesingle s central proposal:
measurement should inform release, but it should not silently become
release authority.

The next priority is to calibrate promotion decisions on larger
independent corpora, add blinded human adjudication, execute real
evaluator and platform integrations, and measure false refusal alongside
false promotion. Until then, the framework's
contribution is a disciplined systems boundary between ``we measured
it'' and ``we are justified in deploying it.''

\appendix
\section{Decision and Evidence Taxonomy}\label{app:taxonomy}

RAGWarrant's evidence taxonomy is deliberately
asymmetric: stronger evidence may support a narrower claim, but weaker
evidence cannot be relabeled upward. The categories below define the
maximum defensible statement associated with each run type.

\begin{longtable}[]{@{}
  >{\raggedright\arraybackslash}p{(\columnwidth - 6\tabcolsep) * \real{0.1857}}
  >{\raggedright\arraybackslash}p{(\columnwidth - 6\tabcolsep) * \real{0.2929}}
  >{\raggedright\arraybackslash}p{(\columnwidth - 6\tabcolsep) * \real{0.2643}}
  >{\raggedright\arraybackslash}p{(\columnwidth - 6\tabcolsep) * \real{0.2571}}@{}}
\caption{Evidence classes and claim ceilings.}\\

\toprule\noalign{}
\begin{minipage}[b]{\linewidth}\raggedright
Evidence class
\end{minipage} & \begin{minipage}[b]{\linewidth}\raggedright
What it establishes
\end{minipage} & \begin{minipage}[b]{\linewidth}\raggedright
Permitted claim example
\end{minipage} & \begin{minipage}[b]{\linewidth}\raggedright
Not permitted
\end{minipage} \\
\midrule\noalign{}
\endhead
\bottomrule\noalign{}
\endlastfoot
Fixture / smoke & Code path, schema, and refusal behavior execute &
Harness runs and artifacts validate. & Benchmark or quality
superiority \\
Development & Candidate behavior on nonsealed data & Promising
development signal. & Confirmatory or production claim \\
Public confirmatory & Held-out result under frozen configuration &
Bounded external signal on this corpus. & Universal generalization \\
Frozen-observation derived & Ablation and counterfactual analysis &
Derived operational comparison. & Independent replication \\
Generative local & Pinned generator and generated-answer scoring &
Bounded local generative validation. & Human or platform validation \\
Human evaluation & Completed blinded annotations & Human-adjudicated
result. & Production outcome without operations \\
Platform-native & Actual platform execution and run records &
Platform-executed benchmark. & Vendor certification \\
Production & Monitored operational outcomes & Production evidence in
declared environment. & General safety guarantee \\
\end{longtable}

\section{Machine-Readable Promotion Decision}\label{app:decision}

The \texttt{promotion\_decision.json} schema separates scientific outcome from
runtime success. A job may execute correctly and still return BLOCK or
INCONCLUSIVE. The record carries the evidence needed for review, CI/CD,
change management, or artifact storage.

\begin{longtable}[]{@{}
  >{\raggedright\arraybackslash}p{(\columnwidth - 4\tabcolsep) * \real{0.1571}}
  >{\raggedright\arraybackslash}p{(\columnwidth - 4\tabcolsep) * \real{0.3714}}
  >{\raggedright\arraybackslash}p{(\columnwidth - 4\tabcolsep) * \real{0.4714}}@{}}
\caption{Promotion-decision schema groups.}\\

\toprule\noalign{}
\begin{minipage}[b]{\linewidth}\raggedright
Field group
\end{minipage} & \begin{minipage}[b]{\linewidth}\raggedright
Representative fields
\end{minipage} & \begin{minipage}[b]{\linewidth}\raggedright
Purpose
\end{minipage} \\
\midrule\noalign{}
\endhead
\bottomrule\noalign{}
\endlastfoot
Identity & run\_id, suite, timestamp\_utc, schema\_version & Trace the
decision to code, config, and evidence. \\
Decision & decision, result\_class, decision\_reason & Separate runtime
status from governance outcome. \\
Selection & selected\_policy, baseline\_policy & Record what was
compared and what, if anything, advances. \\
Effects & quality\_delta, cost\_delta, latency\_delta,
evidence\_support\_delta & Expose the empirical basis. \\
Risk & risk\_flags, claim\_boundaries, validator\_status & Prevent
operational gains from overriding disqualifiers. \\
Artifacts & artifact\_uris, manifest hashes & Support independent review
and tamper detection. \\
\end{longtable}

\section{Current Claim Status}\label{app:claims}

The release candidate supports a bounded governance-and-systems claim.
It does not support optimizer superiority, stable generative efficiency
gains, human validation, official platform benchmarking, production
readiness, or hallucination elimination.

\begin{longtable}[]{@{}
  >{\raggedright\arraybackslash}p{(\columnwidth - 4\tabcolsep) * \real{0.2500}}
  >{\raggedright\arraybackslash}p{(\columnwidth - 4\tabcolsep) * \real{0.2071}}
  >{\raggedright\arraybackslash}p{(\columnwidth - 4\tabcolsep) * \real{0.5429}}@{}}
\caption{Explicitly supported and unsupported claims.}\\

\toprule\noalign{}
\begin{minipage}[b]{\linewidth}\raggedright
Claim
\end{minipage} & \begin{minipage}[b]{\linewidth}\raggedright
Status at v0.1.1-rc1
\end{minipage} & \begin{minipage}[b]{\linewidth}\raggedright
Evidence boundary
\end{minipage} \\
\midrule\noalign{}
\endhead
\bottomrule\noalign{}
\endlastfoot
Open-source governance engine & Supported & Fresh clone, CLI, public
mini, schemas, tests, Docker job \\
CRAG source/retrieval governance signal & Supported with boundaries &
Mock-API result is cost/latency driven and uses a restricted dataset \\
Selector governance blocks unsafe choices & Supported as stress-test
evidence & Packaged cases; not a population estimate \\
RAG Compass superiority & Unsupported & Ranks below alternatives in
multiple runs \\
Stable generative cost/latency superiority & Unsupported & Positive
primary slice did not replicate \\
Human validation & Unsupported & No completed annotations \\
Official platform benchmarking & Unsupported & Adapters/templates only;
no platform-native runs \\
Production readiness & Unsupported & Local hardened Docker validation is
not production operation \\
Hallucination elimination & Unsupported & No such guarantee is tested \\
\end{longtable}

\section{Statistical Audit History}\label{app:audit}

An early offline public real-RAG grouped-split run reported RAG Compass
ahead of a validation-selected baseline by 0.0488 with a zero-width
bootstrap interval. The result appeared precise because 49,850 paired
example-seed rows were available across 4,985 unique examples and 10
seeds.

A row-level reconstruction found only one unique paired delta. Query,
cluster, dataset-blocked, seed, and hierarchical resampling therefore
repeated the same aggregate policy difference rather than exposing
query-level variation. RAGWarrant retained the original run as
descriptive candidate-output evidence but marked it inconclusive for
calibrated query-level uncertainty.

The episode motivated three changes: checks for nonconstant paired
signals, refusal to report one bootstrap under multiple labels, and a
historical evidence ledger that can narrow a prior claim without
deleting the record. It is included here because it illustrates the
framework's correction mechanism rather than a current
empirical result.

\end{document}